\documentclass[11pt,a4paper]{article}
\usepackage[draft]{hyperref} 
\usepackage[hyperref]{acl2018}
\usepackage{times}
\usepackage{latexsym}

\usepackage{url}
\usepackage{times}
\usepackage{xcolor}
\usepackage{soul}
\usepackage[utf8]{inputenc}
\usepackage[small]{caption}
\usepackage{graphicx}  
\usepackage{amssymb}
\usepackage{amsmath}

\usepackage{enumitem}

\renewcommand{\vec}[1]{\mathbf{#1}}
\renewcommand{\matrix}[1]{\mathbf{#1}}

\aclfinalcopy 
\def\aclpaperid{968} 

\title{Generating Fine-Grained
Open Vocabulary Entity Type Descriptions}

\author{Rajarshi Bhowmik \and Gerard de Melo \\
  Department of Computer Science\\
  Rutgers University -- New Brunswick\\
  Piscataway, NJ, USA\\
  {\tt \{rajarshi.bhowmik, gerard.demelo\}@cs.rutgers.edu}
}

\date{}

\begin{document}

\maketitle

\begin{abstract} 
  While large-scale knowledge graphs provide vast amounts of structured facts about entities, 
  a short textual description can often be useful to succinctly characterize an entity and its type. Unfortunately, many knowledge graph entities lack such textual descriptions. In this paper, we introduce a dynamic memory-based network that generates a short open vocabulary description of an entity by jointly leveraging induced fact embeddings as well as the dynamic context of the generated sequence of words. We demonstrate the ability of our architecture to discern relevant information for more accurate generation of type description by pitting the system against several strong baselines.
\end{abstract}

\section{Introduction}
Broad-coverage knowledge graphs such as Freebase, Wikidata, and NELL are increasingly being used in many 
NLP and AI
tasks. For instance, DBpedia and YAGO were vital for IBM's Watson!\ Jeopardy system \cite{Welty2012WatsonAnswerTyping}. Google's Knowledge Graph is tightly integrated into its search engine, yielding improved responses for entity queries as well as for question answering. 
In a similar effort, Apple Inc.\ is building an in-house knowledge graph to power Siri and its next generation of intelligent products and services. 

\begin{figure}[tb]
\centering
\includegraphics[width=0.7\linewidth]{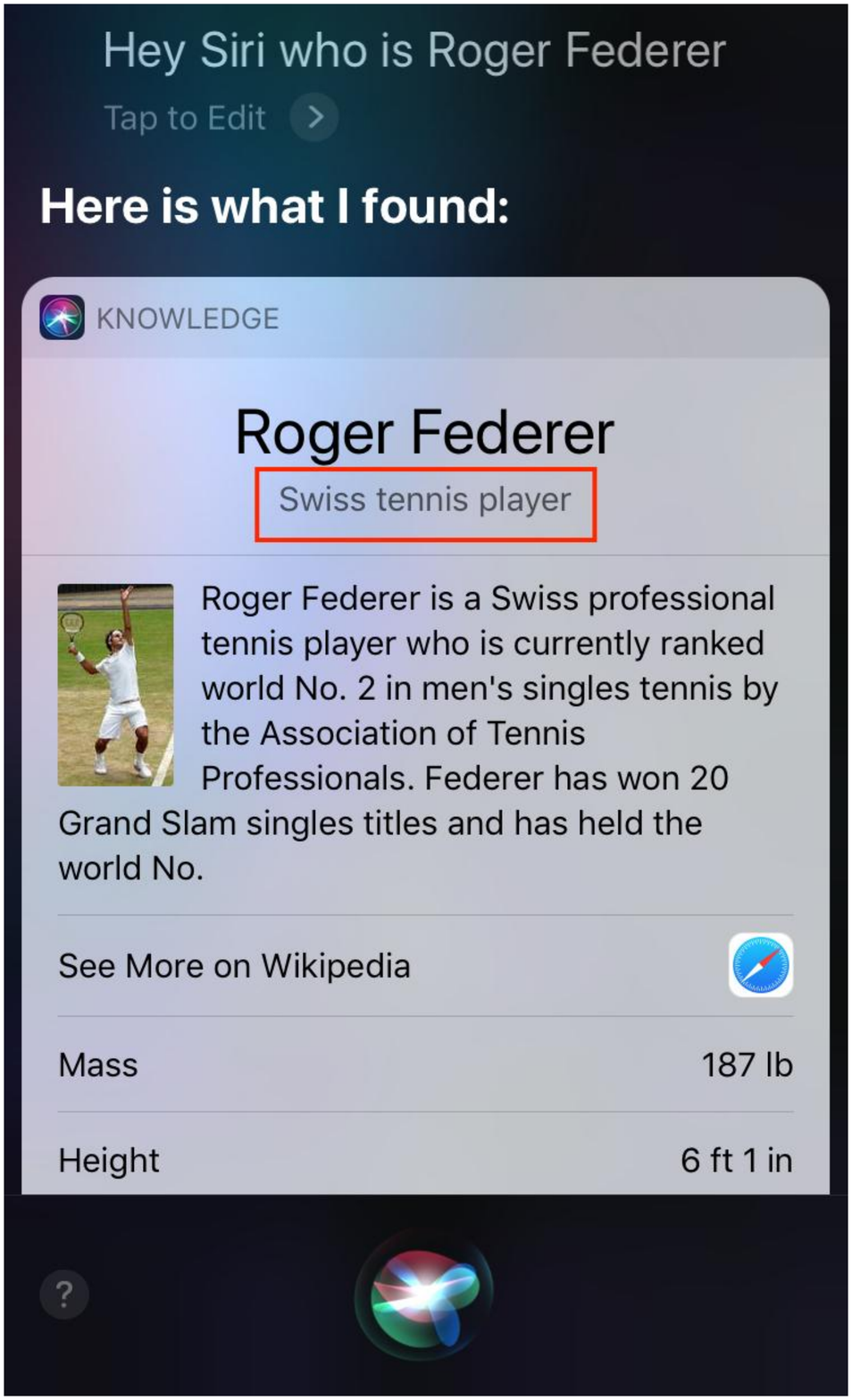}
\caption{A motivating example question that demonstrates the importance of short textual descriptions.}
\label{fig:roger_federer}
\end{figure}

Despite being rich sources of factual knowledge, cross-domain knowledge graphs often lack a succinct textual description for many of the existing entities. Fig.\ \ref{fig:roger_federer} depicts an example of a concise entity description presented to a user. Descriptions of this sort can be beneficial both to humans and in downstream AI and natural language processing tasks, including question answering (e.g., \emph{Who is Roger Federer?}), named entity disambiguation (e.g., \emph{Philadelphia} as a city vs.\ the film or even the brand of cream cheese), and information retrieval, to name but a few. 

Additionally, descriptions of this sort can also be useful to determine the ontological type of an entity -- another challenging task that often needs to be addressed in cross-domain knowledge graphs.
Many knowledge graphs already provide ontological type information, and there has been substantial previous research on how to predict such types automatically for entities in knowledge graphs \cite{neelakantan-chang:2015:NAACL-HLT,MiaoFSZFMS16,KejriwalS17a}, in semi-structured resources such as Wikipedia \cite{Ponzetto2007,deMeloWeikum2010MENTA}, or even in unstructured text \cite{Snow2006,BansalEtAl:2014,TandonEtAl2015Knowlywood}. However, most such work has targeted a fixed inventory of types from a given target ontology, many of which are more abstract in nature (e.g., \emph{human} or \emph{artifact}). In this work, we consider the task of generating more detailed open vocabulary descriptions (e.g., \emph{Swiss tennis player}) that can readily be presented to end users, generated from facts in the knowledge graph. 

Apart from type descriptions, certain knowledge graphs, such as Freebase and DBpedia, also provide a paragraph-length textual abstract for every entity. In the latter case, these are sourced from Wikipedia. There has also been research on generating such abstracts automatically \cite{BiranM17}. While abstracts of this sort provide considerably more detail than ontological types, they are not sufficiently concise to be grasped at a single glance, and thus the onus is put on the reader to comprehend and summarize them.

Typically, a short description of an entity will hence need to be synthesized just by drawing on certain most relevant facts about it. While in many circumstances, humans tend to categorize entities at a level of abstraction commonly referred to as basic level categories \cite{Rosch1976basicobjects}, in an information seeking setting, however, such as in Fig.~\ref{fig:roger_federer}, humans naturally expect more detail from their interlocutor. For example, \emph{occupation} and \emph{nationality} are often the two most relevant properties used in describing a person in Wikidata, while terms such as \emph{person} or \emph{human being} are likely to be perceived as overly unspecific.
However, choosing such most relevant and distinctive attributes from the set of available facts about the entity is non-trivial, especially given the diversity of different kinds of entities in broad-coverage knowledge graphs. Moreover, the generated text should be coherent, succinct, and non-redundant. 

To address this problem, we propose a dynamic memory-based generative network that can generate short textual descriptions from the available factual information about the entities. 
To the best of our knowledge, we are the first to present neural methods to tackle this problem. Previous work has suggested generating short descriptions using predefined templates (cf.\ Section \ref{sec:relatedwork}). However, this approach severely restricts the expressivity of the model and hence such templates are typically only applied to very narrow classes of entities. In contrast, our goal is to design a broad-coverage open domain description generation architecture.

In our experiments, we induce a new benchmark dataset for this task by relying on Wikidata, which has recently emerged as the most popular crowdsourced knowledge base, following Google's designation of Wikidata as the successor to Freebase \cite{FreebaseToWikidata}. With a broad base of 19,000 casual Web users as contributors, Wikidata is a crucial source of machine-readable knowledge in many applications. Unlike DBpedia and Freebase, Wikidata usually contains a very concise description for many of its entities. However, because Wikidata is based on user contributions, many new entries are created that still lack such descriptions. 
This can be a problem for downstream tools and applications using Wikidata for background knowledge.  
Hence, even for Wikidata, there is a need for tools to generate fine-grained type descriptions.
Fortunately, we can rely on the entities for which users have already contributed short descriptions to induce a new benchmark dataset for the task of automatically inducing type descriptions from structured data.

\section{A Dynamic Memory-based Generative Network Architecture}
\begin{figure*}[t]
\centering
\includegraphics[width=0.7\linewidth]{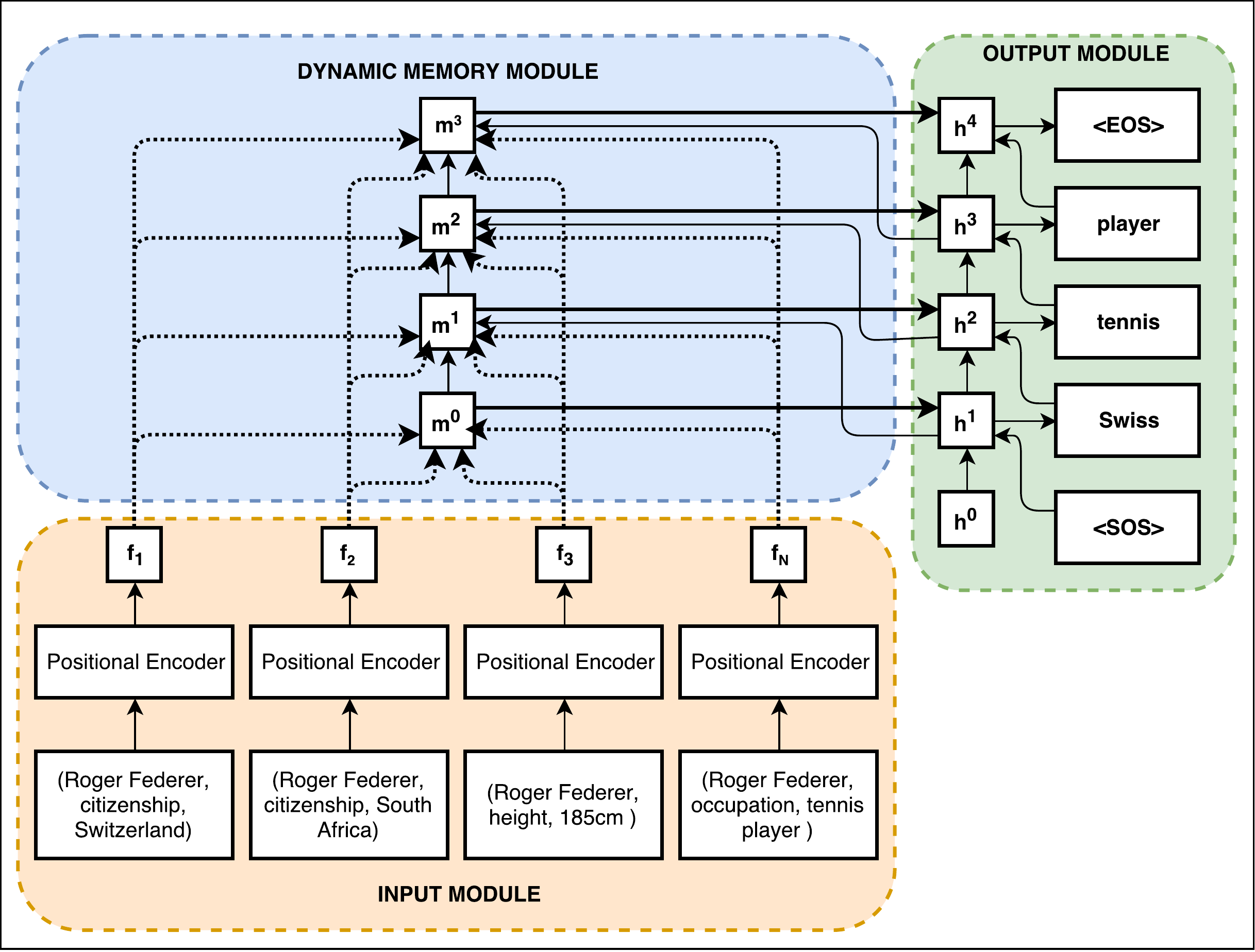}
\caption{Model architecture.}
\label{fig:model_architecture}
\end{figure*}

Our proposed dynamic memory-based generative network
consists of three key components: an input module, a dynamic memory module, and an output module. A schematic diagram of these are given in Fig.~\ref{fig:model_architecture}.

\subsection{Input Module}
\label{sec:model:input}
The input to the input module is a set of $N$ facts $F = \{f_1, f_2,\dots, f_N \}$ pertaining to an entity. Each of these input facts are essentially $(s, p, o)$ triples, for subjects $s$, predicates $p$, and objects $o$. Upon being encoded into a distributed vector representation, we refer to them as \emph{fact embeddings}.

Although many different encoding schemes can be adopted to obtain such fact embeddings, we opt for a positional encoding as described by \newcite{NIPS2015:Sukhbaatar}, motivated in part by the considerations given by \newcite{Xiong:2016:DMN}. For completeness, we describe the positional encoding scheme here. 

We encode each fact $f_i$ as a vector $\vec{f_i} = \sum_{j=1}^J \vec{l_j} \circ \vec{w_j^i}$, where $\circ$ is an element-wise multiplication, and $\vec{l_j}$ is a column vector with the structure $l_{kj} =(1-\frac{j}{J})-(k/d)(1-2\frac{j}{J})$, with $J$ being the number of words in the factual phrase, $ \vec{w_j^i}$ as the embedding of the $j$-th word, and $d$ as the dimensionality of the embedding. Details about how these factual phrases are formed for our data are given in Section \ref{sec:experiment}.

Thus, the output of this module is a concatenation of $N$  fact embeddings $\vec{F} = [\vec{f_1}; \vec{f_2}; \dots; \vec{f_N}]$.

\subsection{Dynamic Memory Module}
The dynamic memory module is responsible for memorizing specific facts about an entity that will be useful for generating the next word in the output description sequence. Intuitively, such a memory should be able to update itself dynamically by accounting not only for the factual embeddings but also for the current context of the generated sequence of words.

To begin with, the memory is initialized as $\vec{m^{(0)}} = \max(\vec{0}, \matrix{W_m}\vec{F} + \vec{b_m})$.
At each time step $t$, the memory module attempts to gather pertinent contextual information by attending to and summing over the fact embeddings in a weighted manner. These attention weights are scalar values informed by two factors: (1) how much information from a particular fact is used by the previous memory state $\vec{m}^{(t-1)}$, and (2) how much information of a particular fact is invoked in the current context of the output sequence $\vec{h}^{(t-1)}$. Formally,
\begin{align}
    \vec{x_i}^{(t)} &= [|\vec{f_i} - \vec{h}^{(t-1)}|;|\vec{f_i} - \vec{m}^{(t-1)}|],\\
    \vec{z_i}^{(t)} &= \matrix{W_2} \tanh({\matrix{W_1}}\vec{x_i}^{(t)} + \vec{b_1}) + \vec{b_2},\\
     a_i^{(t)} &= \frac{\mathrm{exp}(\vec{z_i}^{(t)})}{\sum_{k=1}^{N}\mathrm{exp}(\vec{z_k}^{(t)})},
\end{align}
where $|.|$ is the element-wise absolute difference and $[;]$ denotes the concatenation of vectors.

Having obtained the attention weights, we apply a soft attention mechanism to extract the current context vector at time $t$ as
\begin{equation}
    \vec{c}^{(t)} = \sum_{i=1}^{N}a_i^{(t)} \vec{f_i}.
\end{equation}
This newly obtained context information is then used along with the previous memory state to update the memory state as follows:
\begin{align}
     \vec{C}^{(t)} &= [\vec{m}^{(t-1)};\vec{c}^{(t)};\vec{h}^{(t-1)}  ] \\
     \vec{m}^{(t)} &= \max(\vec{0}, \matrix{W_m} \vec{C}^{(t)} + \vec{b_m}) 
\end{align}
Such updated memory states serve as the input to the decoder sequence of the output module at each time step.

\subsection{Output Module}
The output module governs the process of repeatedly decoding the current memory state so as to emit the next word in an ordered sequence of output words. 
We rely on GRUs for this.

At each time step, the decoder GRU is presented as input a glimpse of the current memory state $\vec{m}^{(t)}$ as well as the previous context of the output sequence, i.e., the previous hidden state of the decoder $\vec{h}^{(t-1)}$. 
At each step, the resulting output of the GRU is concatenated with the context vector $\vec{c_i}^{(t)}$ and is passed through a fully connected layer and finally through a softmax layer. During training, we deploy \emph{teacher forcing} at each step by providing the vector embedding of the previous correct word in the sequence as an additional input. During testing, when such a signal is not available, we use the embedding of the predicted word in the previous step as an additional input to the current step.
Formally,
\begin{align}
    \vec{h}^{(t)} &= \textrm{GRU}([\vec{m}^{(t)}; \vec{w}^{(t-1)}],\vec{h}^{(t-1)}), \\
    \vec{\Tilde{h}}^{(t)} &= \tanh (\matrix{W_d}[\vec{h}^{(t)}; \vec{c}^{(t)}] + \vec{b_d}),\\
    \vec{\hat{y}}^{(t)} &= \textrm{Softmax}(\matrix{W_o}\vec{\Tilde{h}}^{(t)} + \vec{b_o}),
\end{align}
where $[;]$ is the concatenation operator,  $\vec{w}^{(t-1)}$ is vector embedding of the previous word in the sequence, and $\vec{\hat{y}}^{(t)}$ is the probability distribution for the predicted word over the vocabulary at the current step.
\subsection{Loss Function and Training}
\label{sec:model:losstraining}

Training this model amounts to picking suitable values for the model parameters $\theta$, which include the matrices $\matrix{W_1}$, $\matrix{W_2}$, $\matrix{W_m}$, $\matrix{W_d}$, $\matrix{W_o}$ and the corresponding bias terms $\vec{b_1}$, $\vec{b_2}$, $\vec{b_m}$, $\vec{b_d}$, and $\vec{b_o}$ as well as the various transition and output matrices of the GRU.

To this end, if each of the training instances has a description with a maximum of $M$ words, we can rely on the categorical cross-entropy over the entire output sequence as the loss function:
\begin{equation}
\mathcal{L}(\theta)= -\sum_{t=1}^{M}\sum_{j=1}^{|\mathcal{V}|} {y_{j}^{(t)}} \log(\hat{y}_{j}^{(t)}).
\end{equation}
where $y_{j}^{(t)} \in \{0,1\}$ and $|\mathcal{V}|$ is the vocabulary size. 

We train our model end-to-end using Adam as the optimization technique.

\section{Evaluation}
In this section, we describe the process of creating our benchmark dataset as well as the baseline methods and the experimental results.

\begin{table*}[ht]
\centering
  \caption{Automatic evaluation results of different models. For a detailed explanation of the baseline models, please refer to Section \ref{eval:baseline}. The best performing model for each column is highlighted in boldface.}
  \label{tab:results}
  \begin{tabular}{|l|c|c|c|c|c|c|c|}
   \hline
    \textbf{Model}&\textbf{B-1} & \textbf{B-2}& \textbf{B-3}&\textbf{B-4}& \textbf{ROUGE-L}& \textbf{METEOR} & \textbf{CIDEr}\\
    \hline
    Facts-to-seq  & 0.404 & 0.324 & 0.274 & 0.242 & 0.433 & 0.214 & 1.627\\
    Facts-to-seq w. Attention & 0.491 & 0.414 & 0.366 & 0.335 & 0.512 & 0.257 & 2.207\\
    Static Memory & 0.374 & 0.298 & 0.255 & 0.223 & 0.383 & 0.185 & 1.328\\
    DMN+ & 0.281 & 0.234 & 0.236 & 0.234 & 0.275 & 0.139 & 0.912\\
    \hline
    Our Model & \textbf{0.611} & \textbf{0.535} & \textbf{0.485} & \textbf{0.461} & \textbf{0.641} & \textbf{0.353} & \textbf{3.295}\\
    \hline
\end{tabular}
\end{table*}

\subsection{Benchmark Dataset Creation}
For the evaluation of our method, we introduce a novel benchmark dataset that we have extracted from Wikidata and transformed to a suitable format. We rely on the official RDF exports of Wikidata, which are  generated regularly \cite{E+14:WikidataRDF}, specifically, the RDF dump dated 2016-08-01, which consists of 19,768,780 entities with 2,570 distinct properties. A pair of a property and its corresponding value represents a fact about an entity. In Wikidata parlance, such facts are called \emph{statements}. We sample a dataset of 10K entities from Wikidata, and henceforth refer to the resulting dataset as WikiFacts10K. Our sampling method ensures that each entity in Wiki\-Facts\-10K has an English description and at least 5 associated statements. 
We then transform each extracted statement into a phrasal form by concatenating the words of the property name and its value. For example, the (subject, predicate, object) triple (\emph{Roger Federer}, \emph{occupation}, \emph{tennis player}) is transformed to \emph{'occupation tennis player'}. We refer to these phrases as the \emph{factual phrases}, which are embedded as described earlier. We randomly divide this dataset into training, validation, and test sets with a 8:1:1 ratio. 
We have made our code and data available\footnote{https://github.com/kingsaint/Open-vocabulary-entity-type-description} for reproducibility and to facilitate further research in this area.

\subsection{Baselines}
\label{eval:baseline}
We compare our model against an array of baselines of varying complexity. We experiment with some variants of our model as well as several other state-of-the-art models that, although not specifically designed for this setting, can straightforwardly be applied to the task of generating descriptions from factual data. 

\begin{enumerate}
    \item \textbf{Facts-to-sequence Encoder-Decoder Model.} This model is a variant of the standard sequence-to-sequence encoder-decoder architecture described by \newcite{NIPS2014_Sutskever}. However, instead of an input sequence, it here operates on a set of fact embeddings $\{\vec{f_1}, \vec{f_2},\dots ,\vec{f_N}\}$, which are emitted by the positional encoder described in Section \ref{sec:model:input}. We initialize the hidden state of the decoder with a linear transformation of the fact embeddings as 
        $\vec{h}^{(0)} = \matrix{W} \vec{F} + \vec{b}$,
    where $\vec{F} = [\vec{f_1}; \vec{f_2};\dots ;\vec{f_N}]$ is the concatenation of $N$ fact embeddings.
    
    As an alternative, we also experimented with a sequence encoder that takes a separate fact embedding as input at each step and initializes the decoder hidden state with the final hidden state of the encoder. However, this approach did not yield us better results.
    
    \item \textbf{Facts-to-sequence Model with Attention Decoder.} The encoder of this model is identical to the one described above. The difference is in the decoder module that uses an attention mechanism.

    At each time step $t$, the decoder GRU receives a context vector $\vec{c}^{(t)}$ as input, which is an attention weighted sum of the fact embeddings. The attention weights and the context vectors are computed as follows:
    \begin{align}
        \vec{x}^{(t)} &= [\vec{w}^{(t-1)}; \vec{h}^{(t-1)}] \\
         \vec{z}^{(t)} &= \matrix{W}\vec{x}^{(t)} + \vec{b} \\
         \vec{a}^{(t)} &= \mathrm{softmax}(\vec{z}^{(t)})\\
          \vec{c}^{(t)} &= \max(\vec{0}, \sum_{i=1}^{N}a_i^{(t)} \vec{f_i})
    \end{align}
    After obtaining the context vector, it is fed to the GRU as input:
    \begin{equation}
        \vec{h}^{(t)} = \textrm{GRU}([\vec{w}^{(t-1)}; \vec{c}^{(t)}], \vec{h}^{(t-1)})
    \end{equation}
    
    \item \textbf{Static Memory Model.}
    This is a variant of our model in which we do not upgrade the memory dynamically at each time step. Rather, we use the initial memory state as the input to all of the decoder GRU steps. 
    
    \item \textbf{Dynamic Memory Network (DMN+).} 
    We consider the approach proposed by \newcite{Xiong:2016:DMN}, which supersedes \newcite{pmlr-v48-kumar16}. However, some minor modifications are needed to adapt it to our task. Unlike the bAbI dataset, our task does not involve any question. The presence of a question is imperative in DMN+, as it helps to determine the initial state of the episodic memory module. Thus, we prepend an interrogative phrase such as \emph{"Who is"} or \emph{"What is"} to every entity name. The question module of the DMN+ is hence presented with a question such as \emph{"Who is Roger Federer?"} or \emph{"What is Star Wars?"}. Another difference is in the output module. In DMN+, the final memory state is passed through a softmax layer to generate the answer. Since most answers in the bAbI dataset are unigrams, such an approach suffices. However, as our task is to generate a sequence of words as descriptions, we use a GRU-based decoder sequence model, which at each time step receives the final memory state $\vec{m}^{(T)}$ as input to the GRU. We restrict the number of memory update episodes to 3, which is also the preferred number of episodes in the original paper.
\end{enumerate}

\subsection{Experimental Setup}
\label{sec:experiment}
For each entity in the WikiFacts10K dataset, there is a corresponding set of facts expressed as factual phrases as defined earlier. Each factual phrase in turn is encoded as a vector by means of the positional encoding scheme described in Section \ref{sec:model:input}.
Although other variants could be considered, such as LSTMs and GRUs, we apply this standard fact encoding mechanism for our model as well as all our baselines for the sake of uniformity and fair comparison.
Another factor that makes the use of a sequence encoder such as LSTMs or GRUs less suitable is that the set of input facts is essentially unordered without any temporal correlation between facts.

We fixed the dimensionality of the fact embeddings and all hidden states to be 100. The vocabulary size is 29K. Our models and all other baselines are trained for a maximum of 25 epochs with an early stopping criterion and a fixed learning rate of 0.001.

To evaluate the quality of the generated descriptions, we rely on the standard BLEU (B-1, B-2, B-3, B-4), ROUGE-L, METEOR and CIDEr metrics, as implemented by \newcite{sharma2017nlgeval}. Of course, we would be remiss not to point out that these metrics are imperfect.
In general, they tend to be conservative in that they only reward generated descriptions that overlap substantially with the ground truth descriptions given in Wikidata. In reality, it may of course be the case that alternative descriptions are equally appropriate. 
In fact, inspecting the generated descriptions, we found that our method often indeed generates correct alternative descriptions.
For instance, Darius Kaiser is described as a \emph{cyclist}, but one could also describe him as a \emph{German bicycle racer}.
Despite their shortcomings, the aforementioned metrics have generally been found suitable for comparing supervised systems, in that systems with significantly higher scores tend to fare better at learning to reproduce ground truth captions.

\begin{figure*}[t]
\centering
\includegraphics[width=1.2\linewidth,trim={0 7cm 0 5cm},clip]{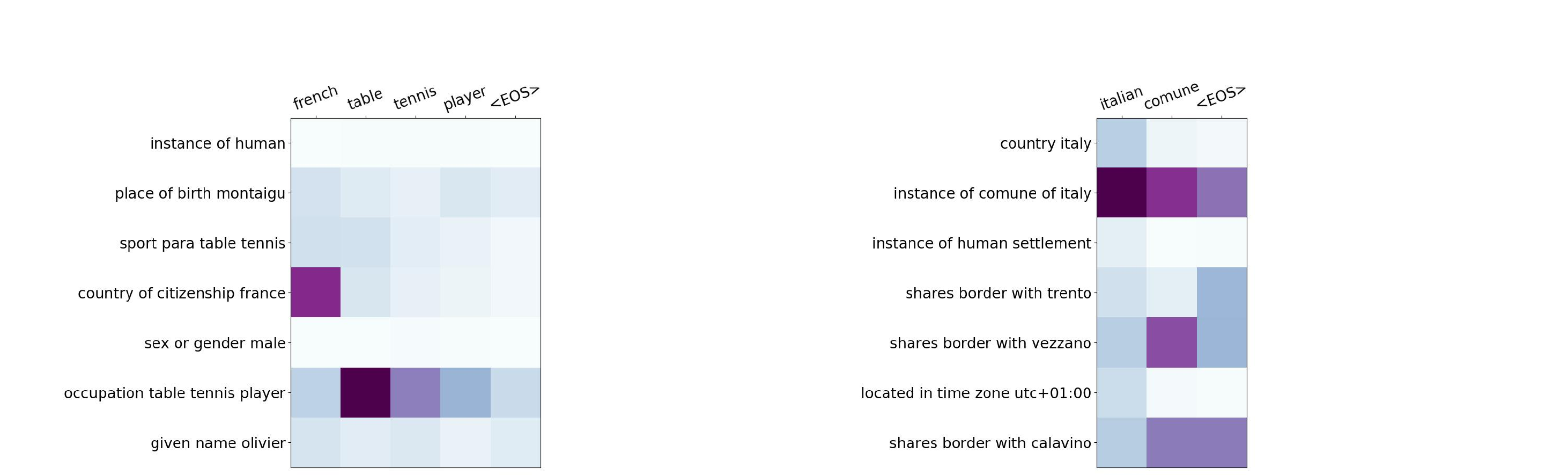}
\caption{An example of attention distribution over the facts while emitting words. The \emph{country of citizenship} property gets the most attention while generating the first word \emph{French} of the left description. For generating the next three words, the fact \emph{occupation} attracts the most attention. Similarly, \emph{instance of} attracts the most attention when generating the sequence \emph{Italian comune}. }
\label{fig:example}
\end{figure*}

\subsection{Results}
The results of the experiments are reported in Table \ref{tab:results}. Across all metrics, we observe that our model obtains significantly better scores than the alternatives.

A facts-to-seq model exploiting our positional fact encoding performs adequately. With an additional attention mechanism (Facts-to-seq w.\ Attention), the results are even better. This is on account of the attention mechanism's ability to reconsider the attention distribution at each time step using the current context of the output sequence. The results suggest that this enables the model to more flexibly focus on the most pertinent parts of the input. In this regard, such a model thus resembles our approach. 
However, there are important differences between this baseline and our model. Our model not only uses the current context of the output sequence, but also memorizes how information of a particular fact has been used thus far, via the dynamic memory module. We conjecture that the dynamic memory module thereby facilitates generating longer description sequences more accurately by better tracking which parts have been attended to,
as is empirically corroborated by the comparably higher BLEU scores for longer n-grams.

The analysis of the Static Memory approach amounts to an ablation study, as it only differs from our full model in lacking memory updates.
The divergence of scores between the two variants suggests that the dynamic memory indeed is vital for more dynamically attending to the facts by taking into account the current context of the output sequence at each step. Our model needs to dynamically achieve different objectives at different time points. For instance, it may start off looking at several  properties to infer a type of the appropriate granularity for the entity (e.g., \emph{village}), while in the following steps it considers a salient property and emits the corresponding named entity for it as well as a suitable preposition (e.g., \emph{in China}).

Finally, the poor results of the DMN+ approach show that a na\"ive application of a state-of-the-art dynamic memory architecture does not suffice to obtain strong results on this task. Indeed, the DMN+ is even outperformed by our Facts-to-seq baseline. This appears to stem from the inability of the model to properly memorize all pertinent facts in its encoder.

\paragraph{Analysis.}
In Figure~\ref{fig:example}, we visualize the attention distribution over facts. We observe how the model shifts its focus to different sorts of properties while generating successive words.

\begin{table*}[ht]
\centering
\small
\caption{A representative sample of the generated descriptions and its comparison with the ground truth descriptions.}
\label{tab:ground_vs_gen}
\begin{tabular}{|l|l|l|l|}
    \hline
    & \textbf{Item} & \textbf{Ground Truth Description} & \textbf{Generated Description} \\
    \hline
    Matches & Q20538915 & painting by Claude Monet & painting by Claude Monet \\
    & Q10592904 & genus of fungi & genus of fungi \\
    & Q669081 & municipality in Austria & municipality in Austria \\
    & Q23588047 & microbial protein found in  & microbial protein found in\\
    &  & ~~~Mycobacterium abscessus & ~~~Mycobacterium abscessus \\
    \hline
    Semantic drift & Q1777131 & album by Hypocrisy & album by Mandy Moore \\
    & Q16164685 & polo player & water polo player\\
    & Q849834 & class of 46 electric locomotives & class of 20 british 0-6-0t locomotives \\
    & Q1434610 & 1928 film & filmmaker \\
    \hline
    More specific & Q1865706 & footballer & Finnish footballer \\
    & Q19261036 & number & natural number \\
    & Q7807066 & cricketer & English cricketer\\
    & Q10311160 & Brazilian lawyer & Brazilian lawyer and politician \\
    \hline
    More general & Q149658 & main-belt asteroid & asteroid \\
    & Q448330 & American musician and pianist & American musician \\
    & Q4801958 & 2011 Hindi film & Indian film \\
    & Q7815530 & South Carolina politician & American politician \\
    \hline
    Alternative & Q7364988 & Dean of York & British academic\\
    & Q1165984 & cyclist & German bicycle racer \\
    & Q6179770 & recipient of the knight's cross & German general \\
    & Q17660616 & singer-songwriter &  Canadian musician \\
    \hline
\end{tabular}
\end{table*}

Table \ref{tab:ground_vs_gen} provides a representative sample of the generated descriptions and their ground truth counterparts. A manual inspection reveals five distinct patterns. The first case is that of exact matches with the reference descriptions. The second involves examples on which there is a high overlap of words between the ground truth and generated descriptions, but the latter as a whole is incorrect because of semantic drift or other challenges. 
In some cases, the model may have never seen a word or named entity during training (e.g., \emph{Hypocrisy}), or their frequency is very limited in the training set. While it has been shown that GRUs with an attention mechanism are capable of learning to copy random strings from the input \cite{gu-EtAl:2016:P16-1},  
we conjecture that a dedicated copy mechanism may help to mitigate this problem, which we will explore in future research. In other cases, the model conflates semantically related concepts, as is evident from  examples such as a \emph{film} being described as a \emph{filmmaker} and a \emph{polo player} as a \emph{water polo player}. 
Next, the third group involves generated descriptions that are more specific than the ground truth, but correct, while, in the fourth group, the generated outputs generalize the descriptions to a certain extent. For example, \emph{American musician and pianist} is generalized as \emph{American musician}, since \emph{musician} is a hypernym of \emph{pianist}.
Finally, the last group consists of cases in which our model generated descriptions that are factually accurate and may be deemed appropriate despite diverging from the reference descriptions to an extent that almost no overlapping words are shared with them. Note that such outputs are heavily penalized by the metrics considered in our evaluation. 

\section{Related Work}\label{sec:relatedwork}

\paragraph{Type Prediction.} There has been extensive work on predicting the ontological types of entities in large knowledge graphs \cite{neelakantan-chang:2015:NAACL-HLT,MiaoFSZFMS16,KejriwalS17a,shimaoka-EtAl:2017:EACLlong}, in semi-structured resources such as Wikipedia \cite{Ponzetto2007,deMeloWeikum2010MENTA}, as well as in text \cite{delcorro-EtAl:2015:EMNLP,yaghoobzadeh-schutze:2015:EMNLP,Ren2016AFETAF}.
However, the major shortcoming of these sorts of methods, including those aiming at more fine-grained typing, 
is that
they assume that the set of candidate types is given as input, and the main remaining challenge is to pick the correct one(s). In contrast, our work yields descriptions that often indicate the type of entity, but typically are more natural-sounding and descriptive (e.g. \emph{French Impressionist artist}) than the oftentimes abstract ontological types (such as \emph{human} or \emph{artifact}) chosen by type prediction methods.

A separate, long-running series of work has obtained open vocabulary type predictions for named entities and concepts mentioned in text \cite{Hearst1992,Snow2006}, possibly also inducing taxonomies from them \cite{Poon:2010:UOI,OntoLearn2013,BansalEtAl:2014}.
However, these methods typically just need to select existing spans of text from the input as the output description.

\paragraph{Text Generation from Structured Data.} Research on methods to generate descriptions for entities has remained scant.
\newcite{LebretGA16} take Wikipedia infobox data as input and train a custom form of neural language model that, conditioned on occurrences of words in the input table, generates biographical sentences as output. However, their system is limited to a single kind of description (biographical sentences) that tend to share a common structure.
\newcite{WangRenTheobaldDyllaDeMelo2016} focus on the problem of temporal ordering of extracted facts.
\newcite{BiranM17} introduced a template-based description generation framework for creating hybrid concept-to-text and text-to-text generation systems that produce descriptions of RDF entities. Their framework can be tuned for new domains,
but does not yield a broad-coverage multi-domain model.
\newcite{ECIR:2017:voskarides}
first create sentence templates for specific entity relationships, and then, given a new relationship instance, generate a description by selecting the best template and filling the template slots with the appropriate entities from the knowledge graph.
\newcite{kutlak} generates referring expressions by converting property-value pairs to text using a hand-crafted mapping scheme.
\newcite{wiseman-shieber-rush:2017:EMNLP2017} considered the related task of mapping tables with numeric basketball statistics to natural language. They investigated an extensive array of current state-of-the-art neural pointer methods but found that template-based models outperform all neural models on this task by a significant margin. However, their method requires specific templates for each domain (for example, basketball games in their case). 
Applying template-based methods to cross-domain knowledge bases is highly challenging, as this would require too many different templates for different types of entities.  
Our dataset contains items of from a large number of diverse domains such as humans, books, films, paintings, music albums, genes, proteins, cities, scientific articles, etc., to name but a few.

\newcite{Chen:2008:LST:1390156.1390173} studied the task of taking representations of observations from a sports simulation (Robocup simulator) as input, e.g.\ \emph{pass(arg1=purple6, arg2=purple3)}, and generating game commentary.
 \newcite{Liang:2009:LSC:1687878.1687893} learned alignments between formal descriptions such as \emph{rainChance(time=26-30,mode=Def)} and natural language weather reports. 
\newcite{mei2016selective} used LSTMs for these sorts of generation tasks, via a custom coarse-to-fine architecture that first determines which input parts to focus on.

Much of the aforementioned work essentially involves aligning small snippets in the input to the relevant parts in the training output and then learning to expand such input snippets into full sentences. In contrast, in our task, alignments between parts of the input and the output do not suffice. Instead, describing an entity often also involves considering all available evidence about that entity to infer information about it that is often not immediately given. Rather than verbalizing facts, our method needs a complex attention mechanism to predict an object's general type and consider the information that is most likely to appear salient to humans from across the entire input.

The WebNLG Challenge \cite{Gardent-EtAl:2017:INLG2017} is another task for generating text from structured data. However, this task
requires a textual verbalization of every triple. On the contrary, the task we consider in this work is quite complementary in that a verbalization of all facts one-by-one is not the sought result. Rather, our task requires synthesizing a short description by carefully selecting the most relevant and distinctive facts from the set of all available facts about the entity. Due to these differences, the WebNLG dataset was not suitable for the research question considered by our paper.

\paragraph{Neural Text Summarization.} Generating entity descriptions is related to the task of text summarization. Most traditional work in this area was extractive in nature, i.e.\ it selects the most salient sentences from a given input text and concatenates them to form a shorter summary or presents them differently to the user \cite{YangDeMelo2017HiText}.
Abstractive summarization goes beyond this in generating new text not necessarily encountered in the input, as is typically necessary in our setting. 
The surge of sequence-to-sequence modeling of text via LSTMs naturally extends to the task of abstractive summarization by training a model to accept a longer sequence as input and learning to generate a shorter compressed sequence as a summary. 

\newcite{rush2015neural} employed this idea to generate a short headline from the first sentence of a text. Subsequent work investigated the use of architectures such as pointer-generator networks to better cope with long input texts \cite{DBLP:conf/acl/SeeLM17}. Recently, \newcite{LiuEtAl2018GeneratingWikipedia} presented a model that generates an entire Wikipedia article via a neural decoder component that performs abstractive summarization of multiple source documents. Our work differs from such previous work in that we do not consider a text sequence as input. Rather, our input are a series of entity relationships or properties, as reflected by our facts-to-sequence baselines in the experiments. 
Note that our task is in certain respects also more difficult than text summarization. 
While regular neural summarizers are often able to identify salient spans of text that can be copied to the output, our input is of a substantially different form than the desired output.

Additionally, our goal is to make our method applicable to any entity with factual information that may not have a corresponding Wikipedia-like article available. Indeed, Wikidata currently has 46 million items, whereas the English Wikipedia has only 5.6 million articles. Hence, for the vast majority of items in Wikidata, no corresponding Wikipedia article is available. In such cases, a summarization baseline will not be effective. 

\paragraph{Episodic Memory Architectures.}
A number of neural models have been put forth that
possess the ability to interact with a memory component.
Recent advances in neural architectures that combine memory components with an attention mechanism exhibit the ability to extract and reason over factual information. 
A well-known example is the End-To-End Memory Network model by \newcite{NIPS2015:Sukhbaatar}, which may make multiple passes over the memory input to facilitate multi-hop reasoning.
These have been particularly successful on the bAbI test suite
of artificial comprehension tests \cite{WestonBCM15Babi}, 
due to their ability to extract and reason over the input.

At the core of the Dynamic Memory Networks (DMN) architecture \cite{pmlr-v48-kumar16} is an episodic memory module, which is updated at each episode with new information that is required to answer a predefined question. Our approach shares several commonalities with DMNs, as it is also endowed with a dynamic memory of this sort. 
However, there are also a number of significant differences.
First of all, DMN and its improved version DMN+ \cite{Xiong:2016:DMN} assume sequential correlations between the sentences and rely on them for reasoning purposes. To this end, DMN+ needs an additional layer of GRUs, which is used to capture sequential correlations among sentences. Our model does not need any such layer, as facts in a knowledge graph do not necessarily possess any sequential interconnections.
Additionally, DMNs assume a predefined number of memory episodes, with the final memory state being passed to the answer module. Unlike DMNs, our model uses the dynamic context of the output sequence to update the memory state. The number of memory updates in our model flexibly depends on the length of the generated sequence.
DMNs also have an additional question module as input, which guides the memory updates and also the output, while our model does not leverage any such guiding factor.
Finally, in DMNs, the output is typically a unigram, whereas our model emits a sequence of words.
\section{Conclusion} 

Short textual descriptions of entities facilitate instantaneous grasping of key information about entities and their types.
Generating them from facts in a knowledge graph requires not only mapping the structured fact information to natural language, but also identifying the type of entity and then discerning the most crucial pieces of information for that particular type from the long list of input facts and compressing them down to a highly succinct form. This is very challenging in light of the very heterogeneous kinds of entities in our data.

To this end, we have introduced a novel dynamic memory-based neural architecture that updates its memory at each step to continually reassess the relevance of potential input signals.
We have shown that our approach outperforms several competitive baselines.
In future work, we hope to explore the potential of this architecture on further kinds of data, including multimodal data \cite{LongChuangDeMelo2018CaptioningMultiFacetedAttention}, from which one can extract structured signals. 
Our code and data is freely available.\footnote{\url{https://github.com/kingsaint/Open-vocabulary-entity-type-description}}

\section*{Acknowledgments}

This research is funded in part by ARO grant no.\ W911NF-17-C-0098 as part of the DARPA SocialSim program.

\bibliography{acl2018}
\bibliographystyle{acl_natbib}
\end{document}